\documentclass[10pt,twocolumn,letterpaper]{article}

\usepackage{cvpr}              

\usepackage{graphicx}
\usepackage{amsmath}
\usepackage{amssymb}
\usepackage{booktabs}

\usepackage[title]{appendix}

\usepackage[pagebackref,breaklinks,colorlinks]{hyperref}

\makeatletter
\newcommand{\printfnsymbol}[1]{%
	\textsuperscript{\@fnsymbol{#1}}%
}
\makeatother

\usepackage[capitalize]{cleveref}
\crefname{section}{Sec.}{Secs.}
\Crefname{section}{Section}{Sections}
\Crefname{table}{Table}{Tables}
\crefname{table}{Tab.}{Tabs.}
\newcommand{\lwb}[1]{\textcolor{black}{#1}}
\newcommand{\lly}[1]{\textcolor{black}{#1}}

\DeclareMathAlphabet{\mathpzc}{OT1}{pzc}{m}{it}
\usepackage{hyperref}
\usepackage{multirow}
\usepackage{multicol}

\begin{document}

\title{Revisiting Temporal Alignment for Video Restoration}

\author{
	Kun Zhou\textsuperscript{1,3}\printfnsymbol{1} \quad Wenbo Li\textsuperscript{2}\thanks{Equal contribution}  \quad Liying Lu\textsuperscript{2} \quad Xiaoguang Han\textsuperscript{1}  \quad Jiangbo Lu\textsuperscript{3}\thanks{Corresponding author}  \\
	\quad ${^1}$The Chinese University of Hong Kong~(Shenzheng), $^{2}$The Chinese University of Hong Kong \\ 
	$^{3}$Smartmore Corporation\\
	{\tt\small kunzhou@link.cuhk.edu.cn},  {\tt\small\{wenboli,lylu\}@cse.cuhk.edu.hk}\\
	{\tt\small\ hanxiaoguang@cuhk.edu.cn},{\tt\small jiangbo.lu@gmail.com} \\
}

\maketitle
\begin{abstract}
Long-range temporal alignment is critical yet challenging for video restoration tasks. Recently, some works attempt to divide the long-range alignment into several sub-alignments and handle them progressively. Although this operation is helpful in modeling distant correspondences, error accumulation is inevitable due to the propagation mechanism.  In this work, we present a novel, generic iterative alignment module which employs a gradual refinement scheme for sub-alignments, yielding more accurate motion compensation. To further enhance the alignment accuracy and temporal consistency, we develop a non-parametric re-weighting method, where the importance of each neighboring frame is adaptively evaluated in a spatial-wise way for aggregation. By virtue of the proposed strategies, our model achieves state-of-the-art performance on multiple benchmarks across a range of video restoration tasks including video super-resolution, denoising and deblurring. Our project is available in \url{https://github.com/redrock303/Revisiting-Temporal-Alignment-for-Video-Restoration.git}.


\end{abstract}

\vspace{-5pt}
\section{Introduction}
\label{intro}
Frame alignment plays an essential role in aggregating temporal information in video restoration tasks, e.g., video super-resolution~(Video SR), video deblurring, and video denoising. In recent years, great attempts have been made to study this problem. Especially, the deep learning-based methods are successful in building temporal correspondences and achieve promising results.

\begin{figure}[t]
	\centering
	
	\includegraphics[width=1.0\columnwidth]{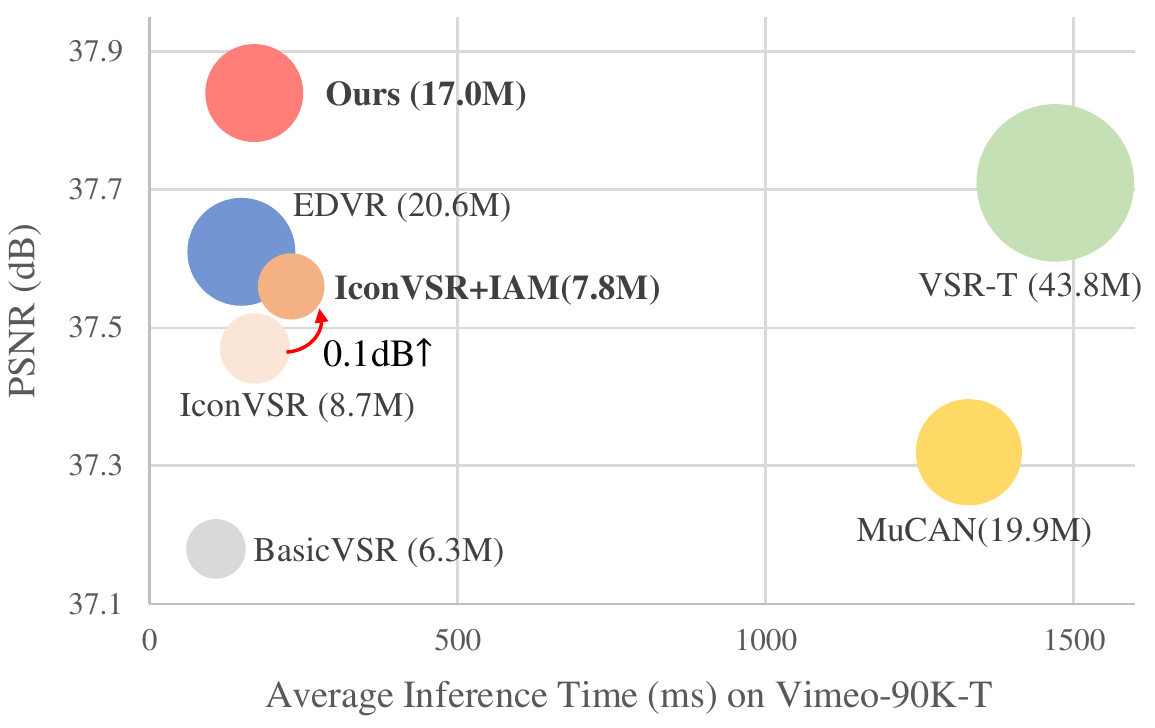}
	\caption{Performance and efficiency comparison on Vimeo-90K-T~\cite{xue2019video}. Besides high PSNR and fast inference, our alignment algorithm can be easily integrated into existing frameworks (e.g., IconVSR~\cite{chan2020basicvsr}) to further improve performance. Circle sizes are set proportional to the numbers of parameters.}
	\label{fig:teasing0}
	\vspace{-8pt}
\end{figure} 
\begin{figure*}[t]
	\centering
	\includegraphics[width=0.95\linewidth]{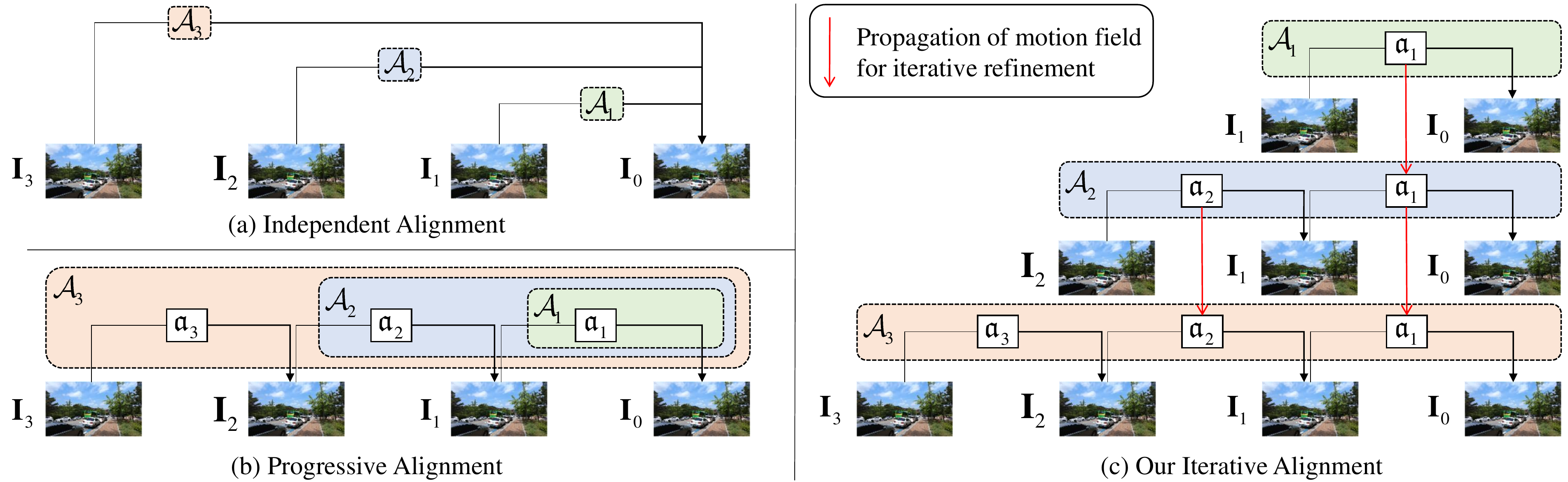} 
	\caption[The LOF caption]{Three alignment strategies in video restoration tasks. (a) Independent alignment that estimates frame-to-frame correspondences in isolation. (b) Progressive alignment that performs multiple alignments sequentially. (c) Our proposed iterative alignment scheme that performs gradual refinement for shared sub-alignments. $\mathcal{A}_{k}$ refers to the $k$-th temporal alignment and $\mathpzc{a}_{i}$ is the $i$-th sub-alignment.  } 
	\label{fig:Teasing}
	\vspace{-0.15in}
\end{figure*}
The existing alignment methods can be roughly categorized into two classes: ($i$) {\it independent alignment} that conducts frame-to-frame alignments totally independently (see~Fig.~\ref{fig:Teasing}(a)) and ($ii$)  {\it progressive alignment} that performs temporally consecutive alignments sequentially in a recursive manner (see~Fig.~\ref{fig:Teasing}(b)). Those independent alignment approaches typically focus on designing effective feature descriptors and motion estimation modules to improve the performance. For example, EDVR~\cite{wang2019edvr} develops pyramid, cascading and deformable convolutions~(PCD) for more accurate alignment. Whereas, without exploiting the correlations between multiple alignments, this strategy is still facing challenges to estimate the long-range motion fields. The second line typically adopts a recurrent framework for gradual alignment. Taking BasicVSR~\cite{chan2020basicvsr} for example, the authors propose an optical-flow-based recurrent architecture for video super-resolution. They predict the bidirectional optical flow between two neighboring frames and then conduct a bidirectional propagation, where the temporal information is aggregated by warping image features produced by \lly{previous steps}. This kind of methods is mainly proposed to model long-range dependencies since it only needs to handle relatively small motion between neighboring frames in one step. However, such chain-rule-based propagation has no chance to correct the misalignment caused by previous steps and may suffer from the error accumulation issue.

As illustrated in Fig.~\ref{fig:Teasing}(c), we observe that different long-range alignments~($A_{i}$) actually share some {\it sub-alignments}~($\mathpzc{a}_{i}$), e.g., $\mathpzc{a}_{1}$ is shared among $A_{1}$, $A_{2}$ and $A_{3}$, so as $\mathpzc{a}_{2}$ in $A_{2}$ and $A_{3}$. How can we utilize this property to improve the accuracy of the shared sub-alignments? In this work, we propose an {\it iterative alignment} module~(IAM) built upon the progressive alignment strategy to gradually refine the shared sub-alignments. For a specific shared sub-alignment (e.g., $\mathpzc{a}_{2}$ in $A_{2}$ and $A_{3}$), the previously estimated result ($\mathpzc{a}_{2}$ in $A_{2}$) is used as a prior in the current iteration ($\mathpzc{a}_{2}$ in $A_{3}$). Our IAM has two merits over the progressive alignment scheme. First, the progressive alignment only conducts a single prediction for each sub-alignment so that misalignment can not be corrected. In contrast, our IAM refines each sub-alignment iteratively, yielding more accurate alignment. Second, the progressive alignment performs multi-frame aggregation based on a chain-like propagation so that misalignment will be propagated to the end. In our IAM, each neighboring frame is aligned through individual propagation, making it more reliable. Furthermore, to reduce the computational complexity, we elaborate a simple yet efficient alignment unit for temporal sub-alignments. From Fig.~\ref{fig:teasing0}, it is observed that our alignment algorithm yields high inference efficiency and superior performance compared with state-the-of-art video SR methods. Particularly, our IAM can be easily plugged into existing deep models. For example, by replacing the original independent alignment module of IconVSR~\cite{chan2020basicvsr} with our ``IAM"~(denoted as ``IconVSR+IAM" in Fig.~\ref{fig:teasing0}), the PSNR is boosted from 37.47dB to 37.56dB on Vimeo-90K-T~\cite{xue2019video}, while reducing the number of parameters from 8.7M to 7.8M.

Besides, the aggregation of multiple aligned frames remains an essential step, for the purpose of preserving details while eliminating alignment errors. Modern restoration systems either employ a sequential of convolutions to directly fuse the aligned features~\cite{tian2020tdan,chan2020basicvsr} or adopt spatial-temporal adaptive aggregation strategies~\cite{wang2019edvr,liu2019image,isobe2020video,li2020learning,wu2020david}. However, all these methods solely rely on the learned parameters, raising the risk of overfitting on a specific domain. In this work, we strive a non-parametric re-weighting module, where two strategies are designed to explicitly evaluate the spatially-adaptive importance of different frames. First, we explore the accuracy of alignments. Patches in the aligned frames are compared with the counterparts in the reference frame, and those of high similarity are assigned with larger weights during fusion. Second, to evaluate the consistency of alignments, we compute the pixel-wise L2 distances of the aligned frames with their average. Pixels with smaller distances are considered to be more consistent with other frames and hence assigned with larger weights. The proposed re-weighting module is parameterless and hence can be plugged into other models.


The main contributions are summarized as:
\begin{itemize}
	\item We rethink issues of the progressive alignment and accordingly propose an iterative alignment scheme, yielding more accurate estimation, especially over long-range correspondences.
	\item We propose a non-parametric re-weighting module that simultaneously evaluates the alignment accuracy and temporal consistency.
	\item The quantitative and qualitative results justify the state-of-the-art performance of our method across several video restoration tasks.
\end{itemize}

\section{Related Work}


{\bf Temporal Alignment.}
Many video restoration approaches~\cite{xue2019video,chan2020basicvsr,wronski2019handheld,pan2020cascaded} perform independent temporal alignment between neighboring frames with the central frame. Various strategies have been proposed to improve the performance. For example, to fill the domain gap between optical flow estimation and video SR tasks, TOF~\cite{xue2019video} integrates a task-oriented flow module into their VSR framework for end-to-end training.  Pan \etal~\cite{pan2020cascaded} develop CNNs to estimate the optical flow and the latent frame simultaneously.  Later on, some methods start to develop adaptive kernel-based schemes~\cite{tian2020tdan,li2020mucan,wang2019edvr,li2020lapar,romano2016raisr,zhou2019kernel,zhou2019spatio,xu2019learning} to perform the alignment and process the occlusion simultaneously.  EDVR~\cite{wang2019edvr} proposes a coarse-to-fine alignment algorithm to tackle the large displacement. However, these independent alignment models only focus on exploring correlations between two frames in isolation. It is still challenging to handle long-range alignments. 

Another line of work~\cite{chan2020basicvsr,chan2021basicvsr++} begins to explore a progressive alignment strategy for video restoration tasks. To alleviate the challenges of long-range alignment, they typically split multiple long-range alignments into several sub-alignments. Those sub-alignments are subsequently processed progressively. In BasicVSR~\cite{chan2020basicvsr}, a pre-trained SPyNet~\cite{ranjan2017optical} is utilized to estimate motion fields of each sub-alignment between adjacent frames. Then, they progressively aggregate the temporal information by warping image features produced by previous steps. The progressive alignment scheme makes it effective in handling long-range alignment. Based on BasicVSR, BasicVSR++~\cite{chan2021basicvsr++} presents a second-order propagation and motion field residual learning method to improve the accuracy of sub-alignments.  However, inaccurately estimated motion fields of some sub-alignments will wrongly warp the image features. The misaligned information is subsequently propagated and aggregated in the later steps, resulting in error accumulation. In this work, we propose an iterative alignment algorithm built upon the progressive alignment scheme. Each sub-alignment is estimated and refined gradually, largely improving the accuracy of alignment.

\begin{figure}[t]
	\centering
	\includegraphics[width=1.0\columnwidth]{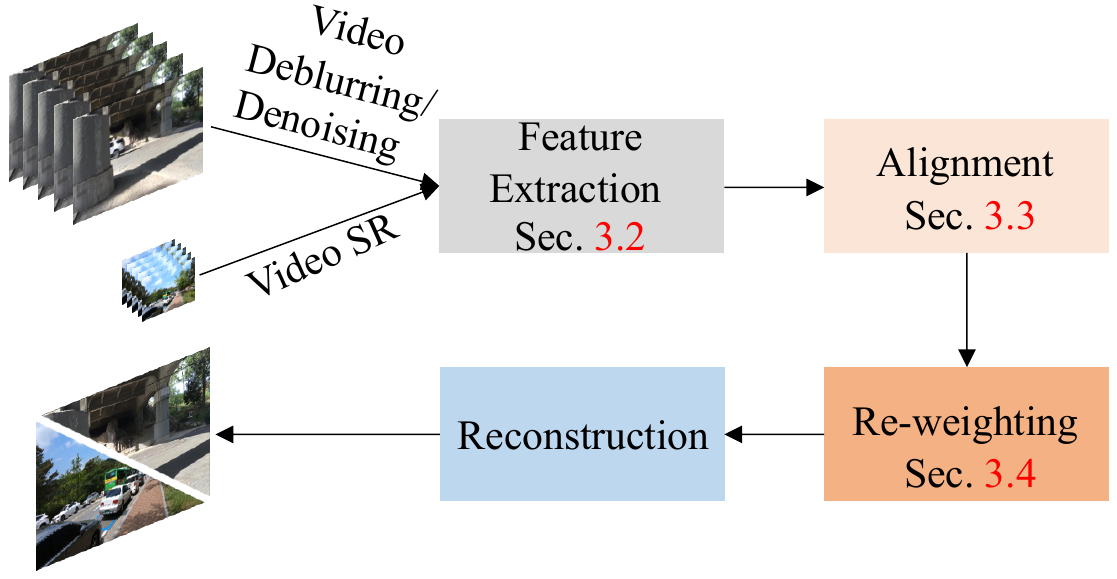} 
	\caption{A general framework for video restoration tasks. There are four components including a frame feature extraction module, an iterative alignment module, an adaptive re-weighting module and a reconstruction module. } 
	\label{fig:framework}
	\vspace{-0.15in}
\end{figure} %


{\bf Feature Fusion.}
The majority of video restoration methods fuse the aligned frames for temporal information aggregation by feature concatenation followed by a convolution~\cite{tian2020tdan,xue2019video,liu2017robust}. For example,
FastDVD~\cite{tassano2020fastdvdnet} divides consecutive frames into different groups and designs a two-stage convolutional neural network for multi-frame fusion. In addition, more effective aggregation strategies have been proposed by applying spatial or temporal attention-based mechanism~\cite{liu2019image,isobe2020video,li2020learning,wu2020david}. Isobe \etal.~\cite{isobe2020video} design a frame-rate-aware group attention, which can handle various levels of motions.  In~\cite{wronski2019handheld}, a motion robustness analysis is adopted to fuse temporal information, where different confidence scores are assigned to the local neighbors of each pixel for merging.
Inspired by this work, we design an adaptive re-weighting module for information aggregation, considering both the accuracy and consistency of the alignment.

\section{Methodology}
\vspace{-3pt}
\subsection{Overview}
\vspace{-5pt}
Fig.~\ref{fig:framework} shows the proposed framework. Our goal is to reconstruct a high-quality image ${\bf I}^{hq}_{0}$ from $2N + 1$ consecutive low-quality images $\{ {\bf I}^{lq}_{-N}, \cdots, {\bf I}^{lq}_{0}, \cdots, {\bf I}^{lq}_{N} \}$. In the feature extraction module, the input frames are first downsampled with strided convolutions for video deblurring/denoising, while being processed under the same resolution for video SR. Then we utilize the proposed IAM to align input frames referring to the central frame. For simplicity, we only consider the one-side alignment in the following as the other side is processed symmetrically. \lly{Afterwards}, an adaptive re-weighting module is designed to fuse the aligned features. Finally, the ${\bf I}^{hq}_{0}$ is obtained by adding the predicted residue to the original (for video deblurring/denoising) or upsampled (for video SR) input image.

\begin{figure}[t]
	\centering
	\includegraphics[width=1.0\columnwidth]{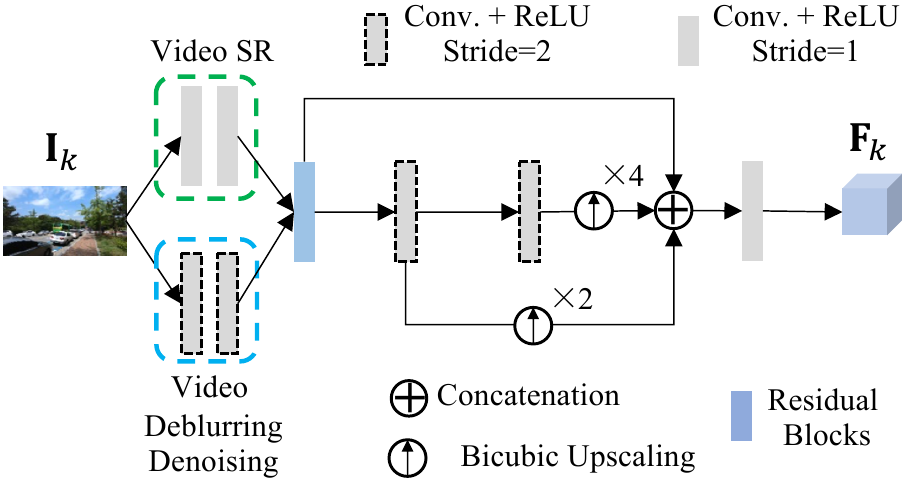} 
	\caption{Overview of our feature extraction module. }
	\label{fig:feaext}
	\vspace{-0.15in}
\end{figure} %

\subsection{Feature Extraction}
\label{sec:fea}
As illustrated in Fig.~\ref{fig:feaext}, we conduct feature extraction to transform a RGB frame ${\bf I}^{lq}_{k}$ to high-dimensional feature maps ${\bf F}_k$. We first utilize two convolutions with strides of 2 to downsample the feature resolutions for video deblurring and denoising~(highlighted in blue \lly{dotted} box in Fig.~\ref{fig:feaext}) for computational efficiency, while keeping the same resolution for video SR~(highlighted in green \lly{dotted} box in Fig.~\ref{fig:feaext}). Then we utilize another two convolutions with \lly{stride} of 2 to obtain the pyramid representations of the input frames. At last, we fuse the pyramid features with a single convolution.
\vspace{-4pt}
\subsection{Temporal Alignment}
Temporal alignment aims to align multiple neighboring features $\{ {\bf F}_{-N}, \cdots, {\bf F}_{-1},{\bf F}_{1},\cdots, {\bf F}_{N} \}$ to a reference ${\bf F}_{0}$. Let $\mathcal{A}_{k}$ be the $k$-th temporal alignment between the neighboring frame ${\bf F}_{k}$ and the reference frame ${\bf F}_{0}$, then we have
\begin{equation}
	\mathcal{A}_{k}( {\bf F}_{k},{\bf F}_{0}) = {\hat{\bf F}}_{k}^0 , \ k\in\{-N,\cdots,-1,1,\cdots,N\},
	\label{eq:a}
\end{equation}
where ${\hat{\bf F}}_{k}^0$ is the aligned result.
\vspace{-8pt}
\subsubsection{Progressive Alignment}
\label{PA}
\vspace{-2pt}

In order to facilitate the long-range alignment, some recent methods~\cite{chan2021basicvsr++} adopt a progressive alignment strategy. For the alignment $\mathcal{A}_{k}$, they \lly{divide it into} sequential sub-alignments $\{ \mathpzc{a}_{k},\mathpzc{a}_{k-1},\cdots,\mathpzc{a}_{1} \}$ to gradually align the feature ${\bf F}_{k}$ to the reference frame ${\bf F}_{0}$. We use $\mathpzc{a}_{i}$ to represent the sub-alginment from ${\bf F}_{i}$ to ${\bf F}_{i-1}$:
\begin{equation}
	\mathpzc{a}_{i}: {\bf F}_{i} \rightarrow {\bf F}_{i-1}.
	\vspace{-4pt}
\end{equation}
As illustrated in Fig.~\ref{fig:Teasing}(b), all neighboring frames are processed through the chained sub-alignments, indicating that the latter sub-alignments strongly depend on the former predictions. Consequently, the error incurred by an intermediate inaccurate sub-alignment will be propagated and accumulated till the end, leading to inferior performance. To alleviate the issue of error accumulation \lly{and boost the restoration quality}, we propose an iterative alignment algorithm to focus on improving the accuracy of each sub-alignment $\mathpzc{a}_{i}$.

\begin{figure}[t]
	\centering
	\includegraphics[width=1.0\columnwidth]{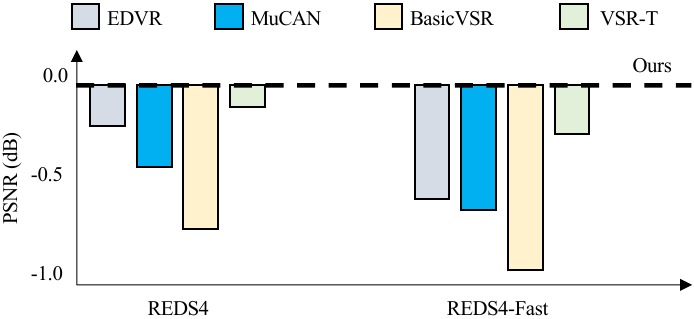} 
	\caption{PSNR differences of four SOTA video SR methods~\cite{wang2019edvr,li2020mucan,chan2020basicvsr,cao2021vsrt} compared to our method (dotted line) on REDS4~\cite{nah2019ntire} and REDS4-Fast.
		The smaller the value, the larger the gap.}
	\label{fig:reds_fast}
	\vspace{-0.15in}
\end{figure} %
\vspace{-9pt}
\subsubsection{Iterative Alignment}
\label{sec:ia}

Unlike the progressive alignment that conducts each sub-alignment only once, our algorithm iteratively refines the sub-alignments based on the previous estimation. As shown in Fig.~\ref{fig:Teasing}(c), we start from the alignment $\mathcal{A}_{1}$, which only contains the sub-alignment $\mathpzc{a}_{1}: {\bf F}_{1} \rightarrow {\bf F}_{0}$, described as:
\begin{equation}
	\mathcal{A}_{1}: \mathpzc{a}_{1}( {\bf F}_{1}, {\bf F}_{0},t=1) \Rightarrow {\hat{\bf F}}_{1}^0,{\bf h}_1^1 \,,
\end{equation}
where ${\hat{\bf F}}_{k}^{i-1}$ refers to the aligned result of sub-alignment $\mathpzc{a}_{i}$ in $\mathcal{A}_{k}$. And ${\bf h}^t_i$ represents the estimated motion field of the sub-alignment $\mathpzc{a}_{i}$ after being refined $t$ times.

After that, we consider the next alignment $\mathcal{A}_{2}$ by sequentially performing two sub-alignments $\{\mathpzc{a}_{2},\mathpzc{a}_{1}\}$:
\begin{equation}
	\mathcal{A}_{2}: 
	\begin{cases}
		\mathpzc{a}_{2}( {\bf F}_{2}, {\bf F}_{1},t=1) \Rightarrow {\hat{\bf F}}_{2}^1,{\bf h}_2^1 \,,  \\
		\mathpzc{a}_{1}( {\hat{\bf F}}_{2}^1, {\bf F}_{0},{\bf h}^1_1,t=2) \Rightarrow {\hat{\bf F}}_{2}^0,{\bf h}_1^2 \,. \\
	\end{cases}
\end{equation}
\lly{For the sub-alignment $\mathpzc{a}_{1}$ in $\mathcal{A}_{2}$, it has already been carried out in $\mathcal{A}_{1}$ once. Thus} we take the pre-estimated motion field ${\bf h}^1_1$ of $\mathpzc{a}_{1}$ in $\mathcal{A}_{1}$ as the initialization and conduct a refinement, formulated as an iterative optimization.

For the subsequent alignment $\mathcal{A}_{3}$, two sub-alignments $\{ \mathpzc{a}_{2}, \mathpzc{a}_{1} \}$ will be refined as:
\begin{equation}
	\mathcal{A}_{3}: 
	\begin{cases}
		\mathpzc{a}_{3}( {\bf F}_{3}, {\bf F}_{2},t=1) \Rightarrow {\hat{\bf F}}_{3}^2,{\bf h}_3^1 \,,  \\
		\mathpzc{a}_{2}( {\hat{\bf F}}_{3}^2, {\bf F}_{1},{\bf h}_2^1,t=2) \Rightarrow {\hat{\bf F}}_{3}^1,{\bf h}_2^2 \,,  \\
		\mathpzc{a}_{1}( {\hat{\bf F}}_{3}^1, {\bf F}_{0},{\bf h}_1^2,t=3) \Rightarrow {\hat{\bf F}}_{3}^0,{\bf h}_1^3 \,.  \\
	\end{cases}
\end{equation}
It can be concluded that, apart from the first sub-alignment $\mathpzc{a}_{k}$ in $\mathcal{A}_{k}$, all other sub-alignments are optimized at least twice. There are two merits: ($i$) The sub-alignments will be more accurate through our iterative refinements. ($ii$) The sub-alignments not only rely on the pre-aligned features but also the pre-estimated motion field, making it more reliable. 

To verify our claim, we evaluate our algorithm together with recent video SR models~\cite{wang2019edvr,chan2020basicvsr,cao2021vsrt,li2020mucan} on REDS4~\cite{nah2019ntire} and REDS4-Fast~\footnote{REDS4-Fast is a subset of REDS4~\cite{nah2019ntire} with an average motion magnitude of {\bf 9.4} pixels, much larger than the average of REDS4 of {\bf 4.3} pixels. The optical flows are calculated by RAFT~\cite{teed2020raft}.}. As shown in Fig.~\ref{fig:reds_fast}, our model achieves the best performance over the competing methods. \lly{Particularly}, our method brings about significant improvement in the context of large motion, demonstrating the effectiveness of our IAM in the long-range alignment.
%
\begin{figure}[t]
	\centering
	\includegraphics[width=1.0\columnwidth]{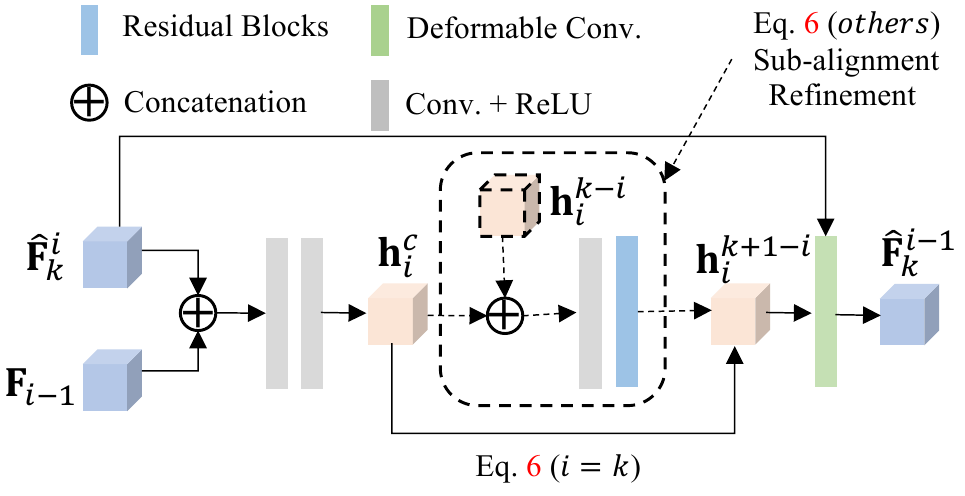} 
	\caption{Illustration of our iterative sub-alignment unit for the sub-alignment $\mathpzc{a}_i$ of $\mathcal{A}_{k}$. ${ \hat{\bf F}}_{k}^i$ is the source feature and ${\bf F}_{i-1}$ is the target feature. The iterative refinement is highlighted in dashed box. ${ \hat{\bf F}}_{k}^{i-1}$ is the aligned result and ${\bf h}_i^{k+1-i}$ is the refined motion field of sub-alignment $\mathpzc{a}_{i}$. }
	\label{fig:framework_align}
	\vspace{-0.15in}
\end{figure} %
 
\vspace{-8pt}
\subsubsection{Sub-Alignment Unit}
In Sec.~\ref{sec:ia}, we describe the iterative alignment algorithm in detail. It is observed that for $2N$ neighboring frames, our method requires $N(N+1)$ sub-alignments. In contrast, the independent and progressive alignment schemes only need $2N$ (sub-)alignments. So it is critical to design a simple sub-alignment unit for computational efficiency. To this end, two improvements have been proposed. ($i$) While the previous methods~\cite{wang2019edvr,chan2020basicvsr} typically adopt a pyramid alignment scheme that performs multiple-scale processing in the alignment phase, we adopt an early multi-scale fusion strategy in the feature extraction phase so that our IAM only performs single-scale alignments. ($ii$) We develop a lightweight sub-alignment unit with much fewer parameters than other methods~\cite{wang2019edvr,cao2021vsrt}. Specifically, we use a compact structure of residual blocks to reduce computational overhead~(see details in supplementary materials.).

Fig.~\ref{fig:framework_align} shows the structure of our sub-alignment unit. Taking the $i$-th sub-alignment $\mathpzc{a}_{i}$ of $\mathcal{A}_{k}$ for example, we first utilize two convolutions followed by ReLU activation to estimate the initialized motion field ${\bf h}_{i}^{c}$ from the concatenation of source feature ${\hat{\bf F}}_{k}^{i}$ and target feature ${\bf F}_{i-1}$.  After that, there are two cases for the prediction ${\bf h}_{i}^{k+1-i}$ of $\mathpzc{a}_{i}$:
\begin{equation}
	{\bf h}_{i}^{k+1-i} = 
	\begin{cases}
		{\bf h}_{i}^{c}, & i=k \,, \\
		\theta({\bf h}_{i}^{c},{\bf h}_{i}^{k-i}), & others \,. \\
	\end{cases}
\end{equation}
If $\mathpzc{a}_{i}$ is the first sub-alignment of $\mathcal{A}_{k}$~($i=k$), then no historical prediction can be reused to refine the $\mathpzc{a}_{i}$. As a result, we simply set ${\bf h}_{i}^{c}$ as the estimated motion field of $\mathpzc{a}_{i}$. Otherwise, we will take the last estimation ${\bf h}_{i}^{k-i}$ together with the current estimation ${\bf h}_{i}^{c}$ as input and utilize a single convolution followed by two residual blocks~(dubbed as $\theta$) to refine the prediction. Finally, we adopt a deformable convolution ~\cite{dai2017deformable} to adaptively sample contents from the source feature ${\hat{\bf F}}_{k}^{i}$:
\begin{equation}
	{\hat{\bf F}}_{k}^{i-1} = {\rm DConv}({\hat{\bf F}}_{k}^{i},{\bf F}_{i-1},{\bf h}_{i}^{k+1-i})
	\label{eq:dcn1}
\end{equation}
Specially, if $\mathpzc{a}_{i}$ is the first sub-alignment in $\mathcal{A}^{k}$ ($i=k$), Eq.~\ref{eq:dcn1} can be written as:
\begin{equation}
	{\hat{\bf F}}_{k}^{k-1} = {\rm DConv}({\bf F}_{k},{\bf F}_{k-1},{\bf h}_{k}^{1})
	\label{eq:dcn2}
\end{equation}
The sub-alignment unit is shared for all sub-alignments, largely reducing the number of learnable parameters.

\subsection{Adaptive Re-weighting}
\label{RMF}

\begin{figure}[t]
	\centering
	\includegraphics[width=1.0\columnwidth]{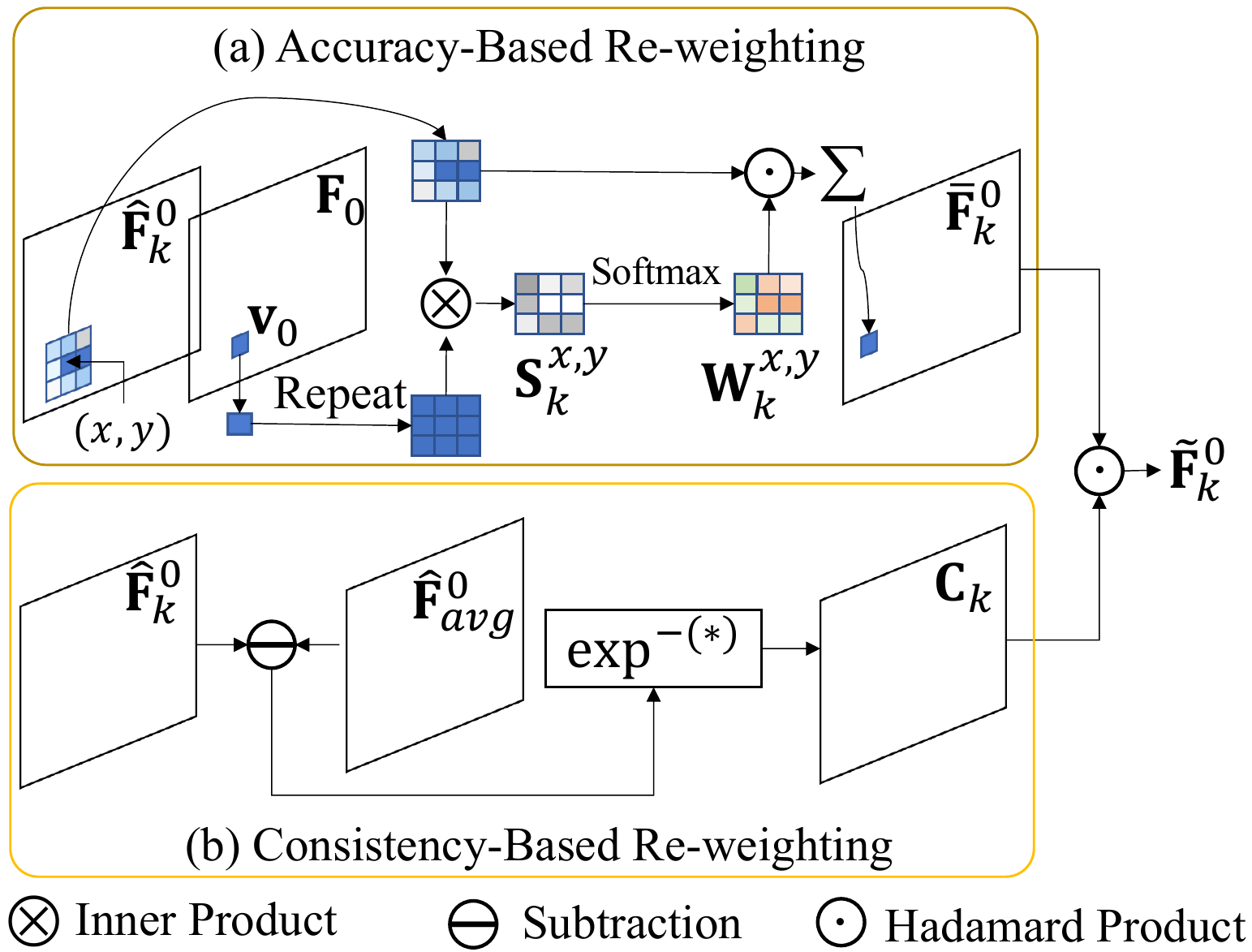}
	\caption{Adaptive re-weighting module. There are two branches: (a) the accuracy-based re-weighting branch for measuring the accuracy of alignment, (b) the consistency-based re-weighting branch for evaluating the consistency of the aligned frames.} 
	\label{fig:rmf}
	\vspace{-0.15in}
\end{figure}

Although the temporal alignment module performs motion compensation for neighboring frames, it remains vital to fuse them in an effective way. Recently, convolution-based attention mechanism becomes popular to aggregate multi-frame information~\cite{liu2019image,isobe2020video,li2020learning,wu2020david}. By contrast, we present a non-parametric re-weighting module to explicitly evaluate the spatially-adaptive importance of aligned frames from two perspectives. First, we evaluate the accuracy of aligned frames with respect to the reference frame. Second, we measure the consistency of aligned neighboring frames. Fig.~\ref{fig:rmf} describes the pipeline of our re-weighting module.
\vspace{-10pt}
\paragraph{Accuracy-Based Re-weighting.} 
As shown in Fig.~\ref{fig:rmf}(a), we measure the accuracy of aligned frames. For the reference frame ${\bf F}_0$, the feature vector at position $(x, y)$ is denoted as ${\bf v}_0$, i.e., ${\bf F}_0(x,y)={\bf v}_0$. We find its corresponding $3 \times 3$ patch centered at the same position in the $k$-th aligned frame $\hat{\bf F}^0_k$. For each feature vector on this patch, we calculate its cosine similarity (normalized inner product) with respect to ${\bf v}_0$ as:
\begin{equation}
	{\bf S}_{k}^{x, y}(\Delta x, \Delta y) = \frac{\hat{\bf F}^{0}_{k}(x+\Delta x,y+\Delta y)}{\left\| \hat{\bf F}^{0}_{k}(x+\Delta x,y+\Delta y) \right\|_{2}} \otimes \frac{{\bf v}_0}{\left\| {\bf v}_0 \right\|_{2}} \,,
\end{equation}
where ${\bf S}_{k}^{x, y}$ is the $3 \times 3$ similarity map at position $(x, y)$ and $\otimes$ represents the inner product. $(x+\Delta x, y+\Delta y)$ is the coordinate of feature vector where $\Delta x, \Delta y \in \left \{ -1, 0, 1 \right \}$. Then a Softmax function is applied to ${\bf S}_k^{x,y}$ in the spatial dimension, yielding the pixel-wise weights as:
\begin{equation}
	{\bf W}_k^{x,y} = \operatorname{Softmax}({\bf S}_k^{x,y}) \,.
\end{equation}
Then ${\bf W}_k^{x,y}$ is used to fuse feature vectors on the $3 \times 3$ patch and the re-weighted result $\bar{\bf F}^0_{k}(x,y)$ is obtained as:
\begin{equation}
	\bar{\bf F}^0_{k}(x,y)=\sum_{\Delta x, \Delta y}{{\bf W}_{k}^{x,y}(\Delta x, \Delta y) \odot \hat{\bf F}^{0}_{k}(x+\Delta x, y+\Delta y)} \,.
\end{equation}
where $\odot$ denotes the Hadamard product.
\vspace{-7pt}
\paragraph{Consistency-Based Re-weighting.} 
We first calculate the average of aligned neighboring frames yielding $\hat{\bf F}^{0}_{avg}$, as illustrated in Fig.~\ref{fig:rmf}(b). For the $k$-th aligned frame $\hat{\bf F}^0_k$, we evaluate its consistency with other aligned frames as
\begin{equation}
	{\bf C}_{k} =  \operatorname{exp}({\alpha \cdot \lVert \hat{\bf F}^{0}_k - \hat{\bf F}^{0}_{avg} \rVert}_2^2) \,,
\end{equation}
where $\alpha$ is set to $-1$ in our experiments. It is noted that ${\bf C}_{k}$ maintains the same shape as $\hat{\bf F}^{0}_k$. \\

Finally, we multiply the accuracy-based re-weighted feature $\bar{\bf F}^0_{k}$ to the consistency map ${\bf C}_{k}$ and obtain the result:
\begin{equation}
	\tilde{\bf F}^{0}_k = \bar{\bf F}^0_{k} \odot {\bf C}_{k} \,.
	\label{mra}
\end{equation}
The refined aligned feature $\tilde{\bf F}^{0}_k$ is passed to the reconstruction module for high-quality image regression~(see Fig.~\ref{fig:framework}).


\section{Experiments}

\subsection{Implementation and Training Details}
\noindent{\bf Configuration.} As shown in Fig.~\ref{fig:framework}, our network consists of four modules: feature extraction, alignment, re-weighting, and reconstruction. The feature extraction module in Sec.~\ref{sec:fea} contains 5 residual blocks for \textit{all} tasks. Table~\ref{table:videoSetting} shows other detailed configurations, where $M$ is the number of feature channels in the network and $B$ is the number of residual blocks in the reconstruction module.
\vspace{0.05in}

\begin{table}[t]
	\small
	\setlength{\tabcolsep}{1pt}
	\centering
	\begin{tabular}{l|c|c|c } 
		\hline
		Task                    &  Video SR                &Video Deblurring          &Video Denoising \\ \hline
		\multirow{2}{*}{Configuration}         &\multirow{2}{*}{$M(128),B(40)$}          & $M(128),B(10)$            & \multirow{2}{*}{$M(64),B(10)$}       \\ 
		                                        &         & $M(128),B(40)$     &            \\ \hline
		GPUs                   & 6                    & 6                      & 2                 \\       
		Patch Reso.          & {$64 \times 64$}       & $128 \times 128$       & $128 \times 128$    \\
		nFrames                & 5(7)                 & 5                     & 5                   \\  
		Mini-Batch          & 4                       & 4                      & 16                   \\  \hline
	\end{tabular}
	\caption{The training and network configurations in different video restoration tasks.  }
	\label{table:videoSetting}
	\vspace{-0.15in}
\end{table}

\noindent{\bf Training.} We show the training settings in Table~\ref{table:videoSetting}. We use 2-6 NVIDIA GeForce RTX 2080 Ti GPUs to train our models for 900K iterations for all three video restoration tasks.  We adopt random vertical or horizontal flipping or $90^{\circ}$ rotation for data augmentation. The initial learning rate is set to $5 \times 10^{-4}$ and a cosine decay strategy is employed. 
\subsection{Datasets and Metrics} 

\noindent{\bf Video Super-Resolution.} REDS~\cite{nah2019ntire} and Vimeo-90K~\cite{xue2019video} are two widely used datasets in Video SR.
Vimeo-90K contains 64,612 training and 7,840 testing 7-frame sequences with resolution $448 \times 256$. The testing set is denoted as Vimeo-90K-T. In REDS, there are 266 training and 4 testing video sequences. Each sequence consists of 100 consecutive frames with resolution $1280 \times 720$. Following~\cite{wang2019edvr}, we denote the testing set as REDS4. Apart from these two testing datasets, we also give the quantitative results on Vid4~\cite{liu2013bayesian}, which consists of 4 video clips.
\vspace{0.05in} 

\noindent{\bf Video Deblurring.} We utilize the video deblurring dataset~\cite{su2017deep} (short for VDB) to train and evaluate our models. There are a total of 61 training and 10 testing video pairs. Each pair contains a blurry and sharp videos. The testing subset is marked as VDB-T.
\vspace{0.05in}

\noindent{\bf Video Denoising.} In this task, we aim to remove Gaussian white noises with known noise levels~($\sigma$). Our model is trained on DAIVS~\cite{khoreva2018video}, which contains 87 training and 30 testing 540p videos. Set8~~\cite{tassano2020fastdvdnet} is also adopted for testing. Following~\cite{tassano2020fastdvdnet}, we keep a maximum of 85 frames for all training and testing sequences for a fair comparison.
\vspace{0.05in}

\noindent{\bf Metrics.} PSNR and SSIM~\cite{wang2004image} are used to evaluate the performance of our models in the video SR and deblurring tasks. For video denoising, following the existing methods~\cite{tassano2020fastdvdnet,chen2016deep}, only PSNR is reported.
\vspace{0.05in}


\begin{table}[t]
	\footnotesize
	\centering
	\setlength{\tabcolsep}{4pt}
	\begin{tabular}{l|c|c|c } 
		\hline
		Task       & Video SR                                    &Video Deblurring          &Video Desnoising  \\ 
		\hline
		Dataset    & Vimeo-90K-T                                 & VDB-T                &DAVIS~($\sigma=20$) \\ \hline
		Baseline              & 37.36                          & 29.88                              & 35.62                 \\       
		+IAM                  & 37.72 (+0.36)                & 32.19 (+2.31)                         & 36.36 (+0.74)                 \\
		+IAM+ARW               & 37.84 (+0.48)                &32.28  (+2.40)                          & 36.73 (+1.11)                  \\  \hline 
	\end{tabular}
	\vspace{4pt}
	\caption{Quantitative comparison for ablation study. PSNR~(dB) is reported. ``Baseline'' means the model without the proposed strategies. ``IAM'' and ``ARW'' denote the iterative alignment module and adaptive re-weighting, respectively.   }
	\label{table:videoAblation}
\end{table}
\begin{figure}[t]
	\centering
	
	\includegraphics[width=1.0\columnwidth]{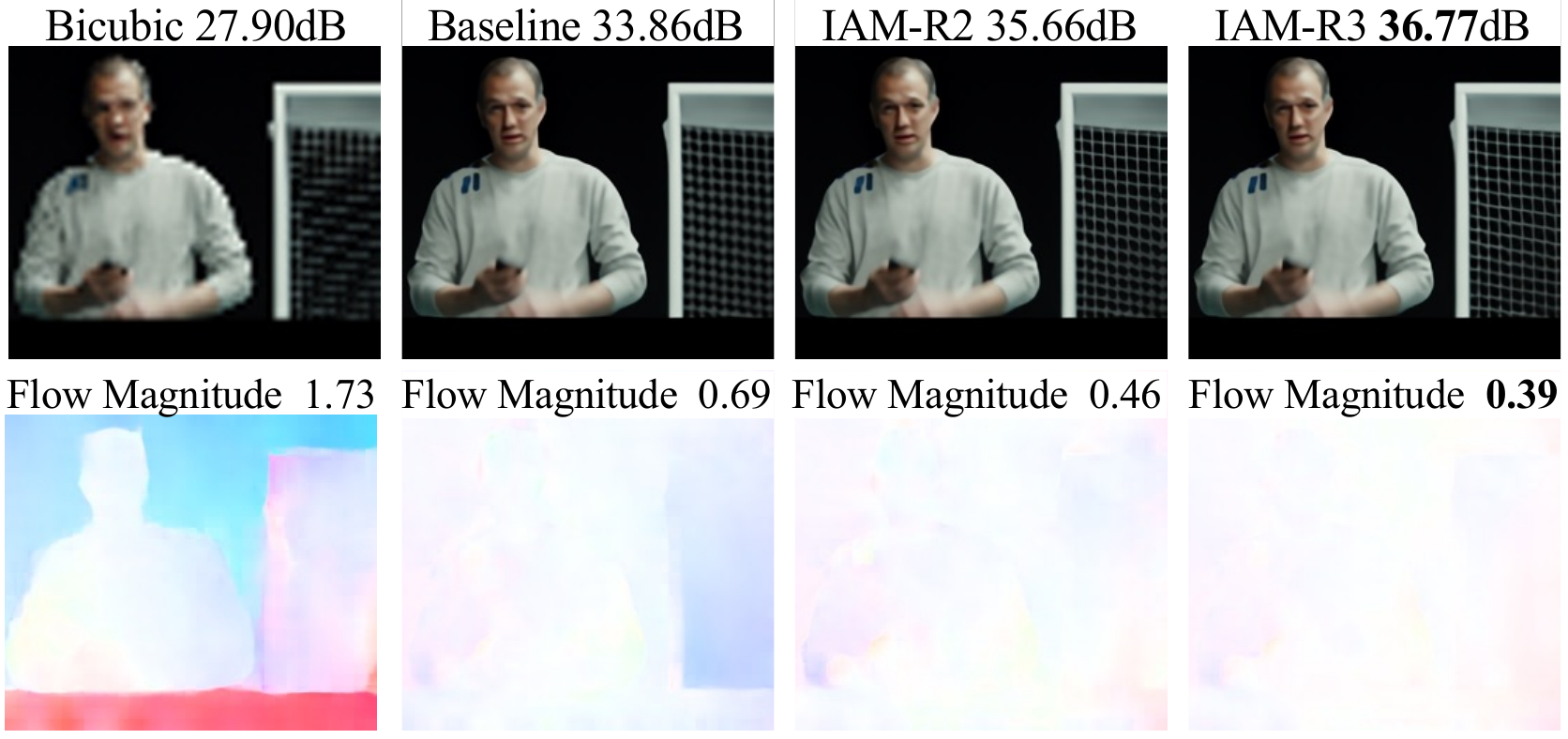}
	\caption{Analysis of the iterative number in IAM in video SR. The first line shows the predictions and the second line shows the optical flows~(using RAFT~\cite{teed2020raft}) between predictions and GT.} 
	\label{fig:viz_iam}
	\vspace{-0.15in}
\end{figure} 

\subsection{Analysis}
In this section, we perform a comprehensive analysis of our method. We abbreviate the iterative alignment module as IAM and the adaptive re-weighting as ARW for clarity.	
\vspace{0.05in}

\noindent{\bf IAM and ARW.} To evaluate the performance of the proposed IAM and ARW designs, we perform a quantitative comparison in Table~\ref{table:videoAblation}. Starting from a baseline without these designs, we incrementally add the iterative alignment module~(IAM) and adaptive re-weighting~(ARW). As illustrated in Table~\ref{table:videoAblation}, the proposed IAM brings about \lwb{0.36dB, 2.31dB and 0.74dB} improvement on PSNR in the video SR, deblurring and denoising tasks, respectively. Besides, we notice that the utilization of ARW further pushes the PSNR up to a new height. Especially, it brings more improvement in the denoising task. All these results manifest the effectiveness of our proposed IAM and ARW strategies.
\vspace{0.05in}


\begin{table}[t]	
	\small
	\setlength{\tabcolsep}{5pt}
	\begin{center}
		\begin{tabular}{c |c|c|c|c}
			\hline
			\multicolumn{2}{c|}{Methods}  & PSNR (dB)  &SSIM & Runtime (ms) \\ 
			\hline
			\multicolumn{2}{c|}{Baseline}  &37.36 &0.9468 &153  \\   \hline
			\multirow{2}{*}{IAM} & R2 & 37.68 (+0.32) &0.9487 &166 \\
			& R3 & 37.72 (+0.36) &0.9490 &169\\
			\hline
			\multirow{2}{*}{ARW} & Acc. & 37.39 (+0.03) &0.9469 &154\\
			& Con. & 37.43 (+0.07) &0.9469 &158\\ \hline
			\multicolumn{2}{c|}{Full}  &37.84 (+0.48) &0.9498 &170  \\    \hline
			
		\end{tabular}
	\end{center}
	\caption{Ablation study on different IAM and ARW settings for video SR. The running time of each model is also reported with an input size of $7\times 112 \times 64$.}
	\label{tab:ablation}
\end{table}

\begin{figure}[t]
	\centering
	\includegraphics[width=1.0\columnwidth]{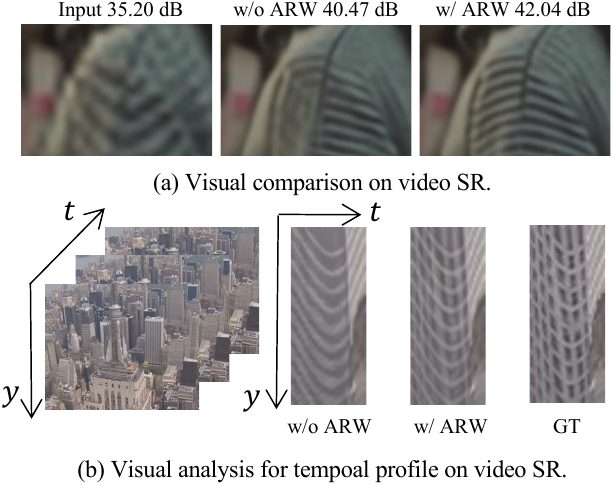}
	\caption{Analysis of the ARW module in Video SR. (a) Visual comparison between without (w/o) and with (w/) ARW. (b) Temporal consistency comparison between without and with ARW.} 
	\vspace{-7pt}
	\label{fig:viz_arw}
\end{figure} 

\noindent{\bf Iterative Number in IAM.}
We assess the influence of iterative number in Table~\ref{tab:ablation} on video SR. Compared to the baseline that performs a single prediction of each sub-alignment~(identical to the progressive alignment), we gradually increase the number of refinements to 2 and 3 (denoted as R2 and R3) for sub-alignments, resulting in PSNR gains by \lwb{0.32dB and 0.36dB}, respectively. It is noteworthy that the increase of running time is acceptable~(13-16ms). Also, as illustrated in Fig.~\ref{fig:viz_iam}, the optical flow between the prediction and GT becomes smaller with the increase of refinements, indicating more accurate alignment. Both quantitative and qualitative results suggest that our IAM can significantly improve the alignment accuracy by reducing the error accumulation during propagation.

\begin{figure}[t]
	\centering
	
	\includegraphics[width=1.0\columnwidth]{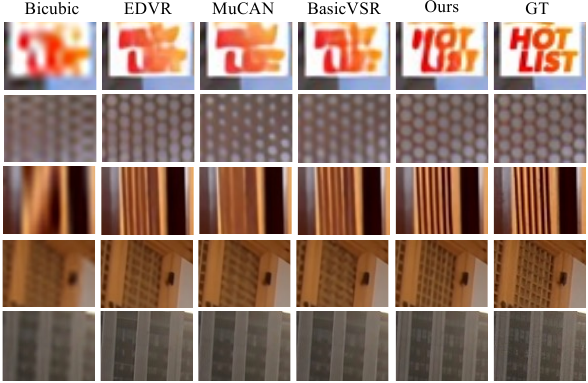}
	\caption{Qualitative comparison on Vimeo-90K-T~\cite{xue2019video} and REDS4~\cite{nah2019ntire} in video SR.} 
	\label{fig:viz_vsr}
\end{figure} 
\noindent{\bf Re-weighting Type in ARW.}
As shown in Table~\ref{tab:ablation}, we study the proposed accuracy- and consistency-based re-weighting strategies for video SR. Compared with the baseline, the accuracy-based re-weighting leads to a \lwb{0.03dB} gain while the consistency one obtains \lwb{0.07dB} improvement, \lly{only costing extra {\bf 1-5ms}}.  Fig.~\ref{fig:viz_arw} shows some examples to illustrate the improved accuracy and consistency of our ARW. It can be observed that the model with our re-weighting module is able to restore more accurate textures while maintaining temporal consistency.

\begin{table}[t]
	
	\small 
	\centering
	\begin{tabular}{l|c|c|c|c } 
	\hline
	\multirow{2}{*}{Method}             			     &\multicolumn{2}{c|}{REDS4 (RGB)}  &\multicolumn{2}{c}{Vid4 (Y)}        \\ \cmidrule(r){2-5} 
	& $N$    & PSNR/SSIM               & $N$    & PSNR/SSIM \\  \hline
	Bicubic                	   	  &1 &26.14/0.7292                   &1  & 23.78/0.6347      \\       
	TOF~\cite{xue2019video}       &5        & 27.98/0.7990           &7  &25.89/0.7651      \\
	DUF~\cite{jo2018deep}         &5       & 28.63/0.8251    		 &7  &27.33/\textbf{{\color{blue}0.8318}} \\
	EDVR~\cite{wang2019edvr}      &5          &31.09/0.8800  		 &7  &27.35/0.8264   \\
	MuCAN~\cite{li2020mucan}      &5         & 30.88/0.8750    		 &7 &27.26/0.8215  \\ 
	VSR-T~\cite{cao2021vsrt}     &5          &{\color {blue}\textbf{31.19/0.8815}}  &7  &27.36/0.8258   \\ 
	{$^\ast$}IconVSR~\cite{chan2020basicvsr}  &5          &30.81/0.8746  &7  &\textbf{{\color{blue}27.39}}/0.8279   \\  \hline
	Ours                          &5       &{\color{red}\textbf{31.30/0.8850}} &7 &{\color {red}\textbf{27.90/0.8380}}  \\ \hline
\end{tabular}
	\caption{REDS4~\cite{nah2019ntire} and Vid4~\cite{liu2013bayesian} results under the $\times 4$ setting in video SR. The PSNR(dB)/SSIM results are evaluated under x4 setting. '$\ast$' indicates the results are from~\cite{cao2021vsrt}. }
	\label{table:VSR_SOTA_REDS}
	\vspace{-0.15in}
\end{table}

\begin{table*}[t]
	
	\setlength{\tabcolsep}{3pt}
	\small 
	\centering
	\begin{tabular}{l|c|c||c|c|c|c|c|c } 
		\hline
		Methods               & Bicubic                    &EDVR~\cite{wang2019edvr}           &MuCAN~\cite{li2020mucan}           &BasicVSR~\cite{chan2020basicvsr}  &IconVSR~\cite{chan2020basicvsr} &$^\dag$BasicVSR++~\cite{chan2021basicvsr++}   &VSR-T~\cite{cao2021vsrt}        &Ours \\  \hline
		nFrame                         &1               &1                                     &7                                           &7                                          &7                               &7                                   &7                                         &7 \\    
		Param.            &-                       &20.6   &19.8               &6.3    &8.7        &7.3     &43.8 &17.0        \\    \hline
		RGB               &29.79/0.8483            &35.79/0.9374   &-     & -        &-  &-  &{\color {blue}\textbf{35.88/0.9380}}  &{\color{red}\textbf{35.96}/\textbf{0.9389}}       \\       
		Y         &31.32/0.8684            &37.61/0.9489   &37.32/0.9465   &37.18/0.9450 &37.47/0.9476     &{\color {blue} \textbf{37.79}}/{\color{red}\textbf{0.9500} } &37.71/0.9494   &{\color{red}\textbf{37.84}}/{\color{blue}\textbf{0.9498}}       \\ \hline 
	\end{tabular}
	\caption{Vimeo-90K-T~\cite{xue2019video} results in video SR. The PSNR(dB)/SSIM results are obtained under the $\times 4$ setting. Numbers in {\color {red} \bf red }and  {\color {blue} \bf blue} refer to the best and second-best results. '$\dag$' means BasicVSR++ uses an {\bf additional REDS dataset } for pretraining. }
	\label{table:VSR_SOTA_VIMEO}
\end{table*}
%
%
%

\begin{table}[t]
	\footnotesize
	\setlength{\tabcolsep}{1pt}
	\centering
	\begin{tabular}{c|c|c|c|c|c|c} 
		\hline
		
		Dataset &  ~ $\sigma$ ~               &VNLB~\cite{arias2018video}                       &V-BM4D~\cite{maggioni2012video}                      &VNLnet~\cite{chen2016deep}           &FastDVD~\cite{tassano2020fastdvdnet}                       &Ours \\  \hline
		\multirow{5}{*}{Set8} & 10                &{\color{red}\textbf{37.26}}                      &36.05                      &37.10                             &36.44                                  &{\color{blue}\textbf{37.25}}         \\       
		& 20                   &33.72                      &32.19                      &{\color{blue}\textbf{33.88}}                             &33.43                                  &{\color{red}\textbf{34.05}}          \\  
		& 30                   &{\color{blue}\textbf{31.74}}                      &30.00                      &-                                 &31.68                                  &{\color{red}\textbf{32.19}}          \\  
		& 40                   &30.39                      &28.48                      &{\color{blue}\textbf{30.55}}                             &30.46                                  &{\color{red}\textbf{30.89}}         \\  
		& 50                   &29.24                      &27.33                      &29.47                             &{\color{blue}\textbf{29.53}}                                  &{\color{red}\textbf{29.90}}   \\  \hline 
		
		\multirow{5}{*}{DAVIS}  & 10                &{\color{blue}\textbf{38.85}}                      &37.58                      &35.83                             &38.71                                  &{\color{red}\textbf{39.75}}          \\       
		& 20                &35.68                      &33.88                      &34.49                             &{\color{blue}\textbf{35.77}}                                   &{\color{red}\textbf{36.73}}         \\  
		& 30                &33.73                      &31.65                      &-                                 &{\color{blue}\textbf{34.04}}                                  &{\color{red}\textbf{34.89}}         \\  
		& 40                &32.32                      &30.05                      &32.32                             &{\color{blue}\textbf{32.82}}                                  &{\color{red}\textbf{33.56}}          \\  
		& 50                &31.13                      &28.80                      &31.43                             &{\color{blue}\textbf{31.86}}                                  &{\color{red}\textbf{32.51}}          \\\hline 
	\end{tabular}
	\caption{Set8~\cite{tassano2020fastdvdnet} and DAVIS~\cite{khoreva2018video} results in video denoising. PSNR(dB) results are reported.} 
	\vspace{-4pt}
	\label{table:DENOISE_SOTA}
\end{table}

\subsection{Comparison with State-of-The-Art Methods}
We compare our method with state-of-the-art approaches quantitatively and qualitatively in the video SR, video deblurring and video denoising tasks.
\vspace{0.05in}






%
\noindent{\bf Video Super-resolution.}
Table~\ref{table:VSR_SOTA_VIMEO} and Table~\ref{table:VSR_SOTA_REDS} exhibit the quantitative results of our method and existing video SR methods~\cite{xue2019video,jo2018deep,wang2019edvr,li2020mucan,chan2020basicvsr,haris2019recurrent,cao2021vsrt,chan2021basicvsr++} on Vimeo-90K-T~\cite{xue2019video}, REDS4~\cite{nah2019ntire} and Vid4~\cite{liu2013bayesian} datasets. Compared to the representative independent~\cite{wang2019edvr} and progressive~\cite{chan2020basicvsr} alignment methods, our method obtains superior Y-channel PSNR performance with 0.23dB and 0.66dB improvement on Vimeo-90K-T, respectively. In addition, our model also surpasses the VSR-T~\cite{cao2021vsrt} by 0.13dB, which has much more parameters. While BasicVSR++~\cite{chan2021basicvsr++} uses an additional dataset for pretraining, our results are still the best. In terms of the Vid4~\cite{liu2013bayesian} dataset, our method achieves significant improvement with {\bf 0.51dB} on PSNR compared with IconVSR~\cite{chan2020basicvsr}. Besides, Fig.~\ref{fig:viz_vsr} shows the visual comparison on Vimeo-90K-T and REDS4. Our model recovers much clearer text and more accurate structures compared to other methods. 
\vspace{0.05in}

%

\begin{figure}[t]
	\centering
	\includegraphics[width=1.0\columnwidth]{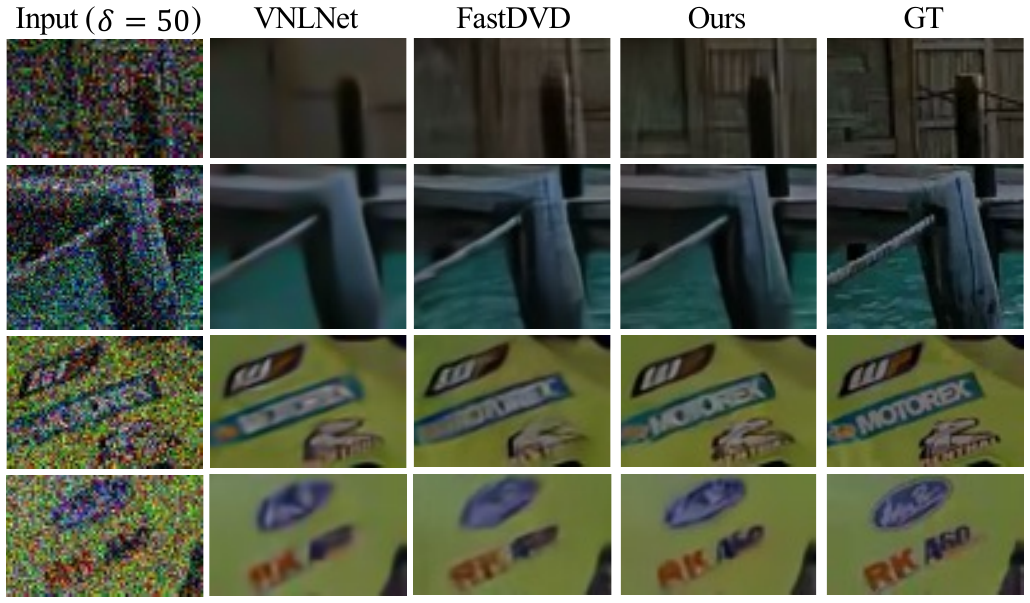} 
	\caption{Qualitative comparison on Set8~\cite{tassano2020fastdvdnet} in video denoising.  } 
	\vspace{-3pt}
	\label{fig:denoisesota}
\end{figure}

\vspace{-2pt}
\noindent{\bf Video Denoising.}
Following previous methods~\cite{tassano2020fastdvdnet,arias2018video,chen2016deep}, we adopt Set8~\cite{tassano2020fastdvdnet} and DAVIS~\cite{khoreva2018video} as our benchmarks in the video denoising task. The quantitative results are reported in Table~\ref{table:DENOISE_SOTA}. Our model achieves the best results under most noise levels. Especially, compared with the second-best approaches, our method largely improves the PSNR by \textbf{0.37dB} and \textbf{0.65dB} under the noise level $\sigma=50$ on Set8 and DAVIS, respectively. Fig.~\ref{fig:denoisesota} presents some qualitative results. It is observed that our method restores richer and clearer textures compared with other approaches.

\begin{figure}[t]
	\centering
	
	\includegraphics[width=1.0\columnwidth]{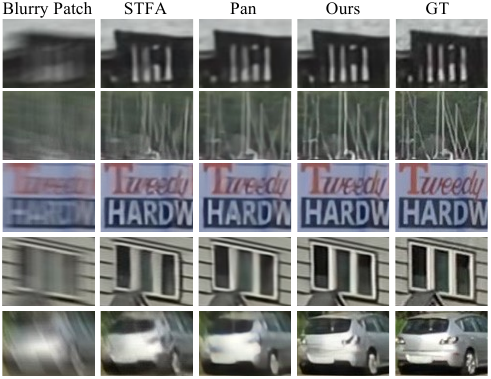}
	\caption{Visual results on VDB-T~\cite{su2017deep} in video deblurring.} 
	\vspace{-3pt}
	\label{fig:viz_blur}
\end{figure}

\vspace{0.05in}
\vspace{-2pt}
\noindent{\bf Video Deblurring.}
We compare our method with several recent video deblurring approaches~\cite{hyun2017online,su2017deep,wang2019edvr,zhou2019spatio,pan2020cascaded} on VDB-T~\cite{su2017deep}. As illustrated in Table~\ref{table:videoSetting}, two models with different sizes (10 or 40 residual blocks) are developed, denoted as ``Ours-M" and ``Ours". From Table~\ref{table:BLUR_SOTA}, compared to the second best ARVo~\cite{pan2020cascaded}, we see that our model achieves {\bf 0.12dB} and 0.013 improvement on PSNR and SSIM, respectively. Some visual examples illustrated in  Fig.~\ref{fig:viz_blur} also demonstrate that our model is able to handle soame challenging cases with complex motion blur.
\vspace{0.05in}

\begin{table}[t]
	\footnotesize
	\setlength{\tabcolsep}{2pt}
	\centering
	\begin{tabular}{l|c|c|c|c|c|c } 
		\hline
		Meth.                       &EDVR~\cite{wang2019edvr}           &STFA~\cite{zhou2019spatio}  &Pan~\cite{pan2020cascaded}  &ARVo~\cite{li2021arvo}        &Ours-M &Ours\\  \hline
		Param.                           &23.6M                            &5.4M                 &16.2M       &-      &12.7M    &16.7M         \\     \hline   
		PSNR                                                        &28.51                             &31.24         &32.13  &{\color{blue}\textbf{32.80}}       &  32.28   &{\color{red}\textbf{32.92}}        \\       
		SSIM                                                         &0.864                             &0.934         &0.927   &0.935             &{\color{blue}\textbf{0.942}}            &{\color{red}\textbf{0.948}}     \\ \hline 
	\end{tabular}
	\vspace{4pt}
	\caption{VDB-T~\cite{su2017deep} results in video deblurring. ``Ours-M'' and ``Ours'' denote our medium and standard models.}
	\label{table:BLUR_SOTA}
	\vspace{-0.15in}
\end{table}

%
%

\vspace{-2pt}

\subsection{Limitation}
In this work, the proposed designs are mainly for improving the accuracy of the long-range alignment. There remains plenty of room to optimize the modeling of subtle motion. Besides, further improving the efficiency of the alignment pipeline is also our future target.


\section{Conclusion}
In this paper, we propose a simple yet effective iterative alignment algorithm (IAM) and an efficient adaptive reweighting strategy~(ARW) to better utilize multi-frame information. The quantitative and qualitative results of three video restoration tasks illustrate the effectiveness of our method. Besides, we show that our method is general to be deployed in existing video processing systems to further improve their performance. We will explore more video-based tasks in the future. The code will be publicly available to promote the development of the community.

{\small
\bibliographystyle{ieee_fullname}
\bibliography{RTA_ARXIV}
}
\begin{appendices}
	\section{Residual Block}
	As illustrated in Table~\ref{tab:residual}, our residual block is comprised of two convolutions, where the first convolution is followed by ReLU activation. In previous works~\cite{wang2019edvr,chan2020basicvsr,cao2021vsrt}, the channel numbers typically keep identical within the residual block. In contrast, we set the channel dimension of hidden representations~($M_1$) no more than 64 to reduce the parameters of our network \lly{and speed up} the training and inference phases:
	\begin{equation}
		M_1 = max(M//2,64),
	\end{equation}
	\vspace{-4pt}
	\begin{table}[h]	
		\begin{center}
			\begin{tabular}{|c|l|c|c}
				\hline
				Input        	& x \\ \hline
				Layer1       	& Conv($M$,$M_1$,3,1) + ReLU \\ \hline
				Layer2       	& Conv($M_1$,$M$,3,1) $\Rightarrow y$\\ \hline 
				Output       	& x+y   \\ \hline 
			\end{tabular}
		\end{center}
		\caption{The structure of our residual block. $M$ refers to the channel number of input.   }
		\label{tab:residual}
		\vspace{-10pt}
	\end{table}
\section{More Reuslts}

\subsection{Temporal Consistency}
\begin{figure}[t]
	\centering
	\includegraphics[width=1.0\columnwidth]{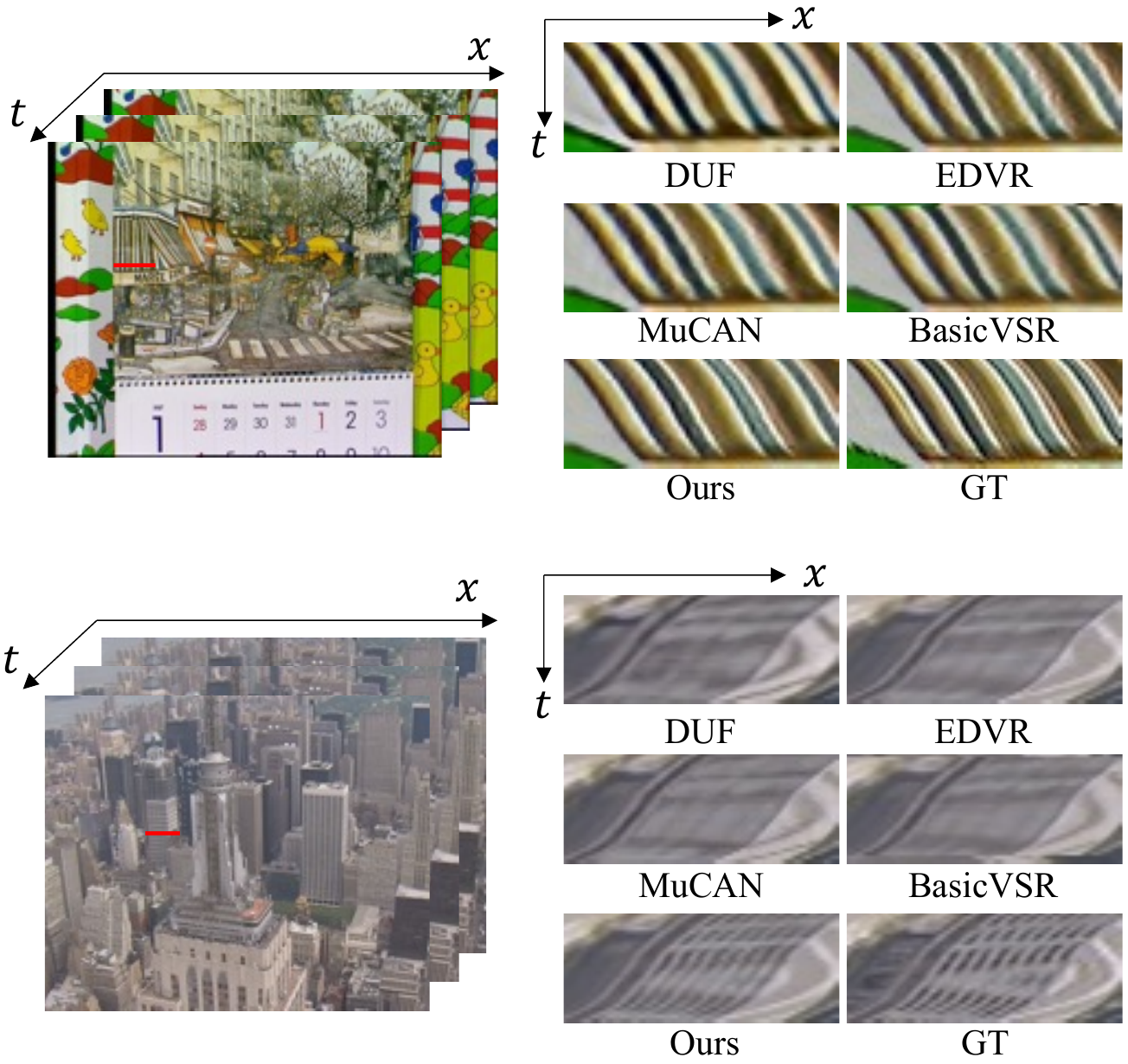} 
	\caption{Visualization of temporal consistency on Vid4~\cite{liu2013bayesian}. }
	\label{fig:temporalcon}
	\vspace{-0.15in}
\end{figure}

We \lly{compare} the temporal consistency of our method with several state-of-art video SR approaches~\cite{jo2018deep,wang2019edvr,li2020mucan,chan2020basicvsr}. The visual results are illustrated in Figure~\ref{fig:temporalcon}. \lly{It is observed that} other methods fail to restore the consistent textures clearly. While our method empowered with iterative alignment and two adaptively reweighting strategies is able to generate realistic image contents that are closest to the \lly{ground-truth}.

\begin{table*}[h]
	\begin{center}
		\small 
		\centering
		\begin{tabular}{l|c|c||c|c|c|c } 
			\hline
			Methods             	&nFrame			 &Clip\_000           &Clip\_011          &Clip\_015                     &Clip\_020             &Average         \\   \hline
			
			Bicubic                		&1   	&24.55/0.6489       &26.06/0.7261       &28.52/0.8034                &25.41/0.7386       &26.14/0.7292           \\       
			RCAN~\cite{zhang2018image} &1      &26.17/0.7371        &29.34/0.8255       &31.85/0.8881               &27.74/0.8293        & 28.78/0.8200      \\ 
			TOF~\cite{xue2019video}    &7     &26.52/0.7540        &27.80/0.7858       &30.67/0.8609               &26.92/0.7953        & 27.98/0.7990      \\
			DUF~\cite{jo2018deep}      &7     &27.30/0.7937        &28.38/0.8056       &31.55/0.8846               &27.30/0.8164        & 28.63/0.8251      \\
			EDVR~\cite{wang2019edvr}    &5     &28.01/0.8250         &32.17/0.8864       &34.06/\textbf{\color{blue}0.9206}            &30.09/0.8881         &31.09/0.8800     \\
			MuCAN~\cite{li2020mucan}   &5     &27.99/0.8219       &31.84/0.8801       &33.90/0.9170               &29.78/0.8811        & 30.88/0.8750      \\  
			{$^\ast$}BasicVSR~\cite{chan2020basicvsr}   &5     &27.67/0.8114       &31.27/0.8740       &33.58/0.9135               &29.71/0.8803        & 30.56/0.8698      \\ 
			{$^\ast$}IconVSR~\cite{chan2020basicvsr}   &5     &27.83/0.8182       &31.69/0.8798       &33.81/0.9164               &29.90/0.8841        & 30.81/0.8746      \\
			VSR-T~\cite{cao2021vsrt}   &5     &\textbf{{\color{blue}28.06}/{\color{blue}0.8267}}      &\textbf{{\color{red}32.28}/{\color{blue}0.8883}}       &\textbf{ {\color{blue}34.15}}/0.9199               &\textbf{{\color{red}30.26}/{\color{blue}0.8912}}        & \textbf{{\color{blue}31.19}/{\color{blue}0.8815}} \\ \hline
			
			Ours                       &5    &\textbf{{\color{red}28.16}/{\color{red}0.8316}}      &\textbf{{\color{blue}32.24}/{\color{red}0.8889}}       &\textbf{ {\color{red}34.53}/{\color{red}0.9275}}               &\textbf{{\color{red}30.26}/{\color{red}0.8920}}        & \textbf{{\color{red}31.30}/{\color{red}0.8850}}   \\ \hline 
			
		\end{tabular}
	\end{center}
	\caption{Quantitative comparison on REDS4~\cite{nah2019ntire} benchmark under $\times 4$ setting for video super-resolution.  Numbers in {\color {red} \bf red }and  {\color {blue} \bf blue} refer to the best and second-best results. All the results are evaluated in the RGB channel. '$\ast$' indicates the results are from~\cite{cao2021vsrt}.}
	\label{table:VSR_SOTA_REDS}
\end{table*}

\begin{figure*}[t]
	\centering
	\includegraphics[width=0.95\linewidth]{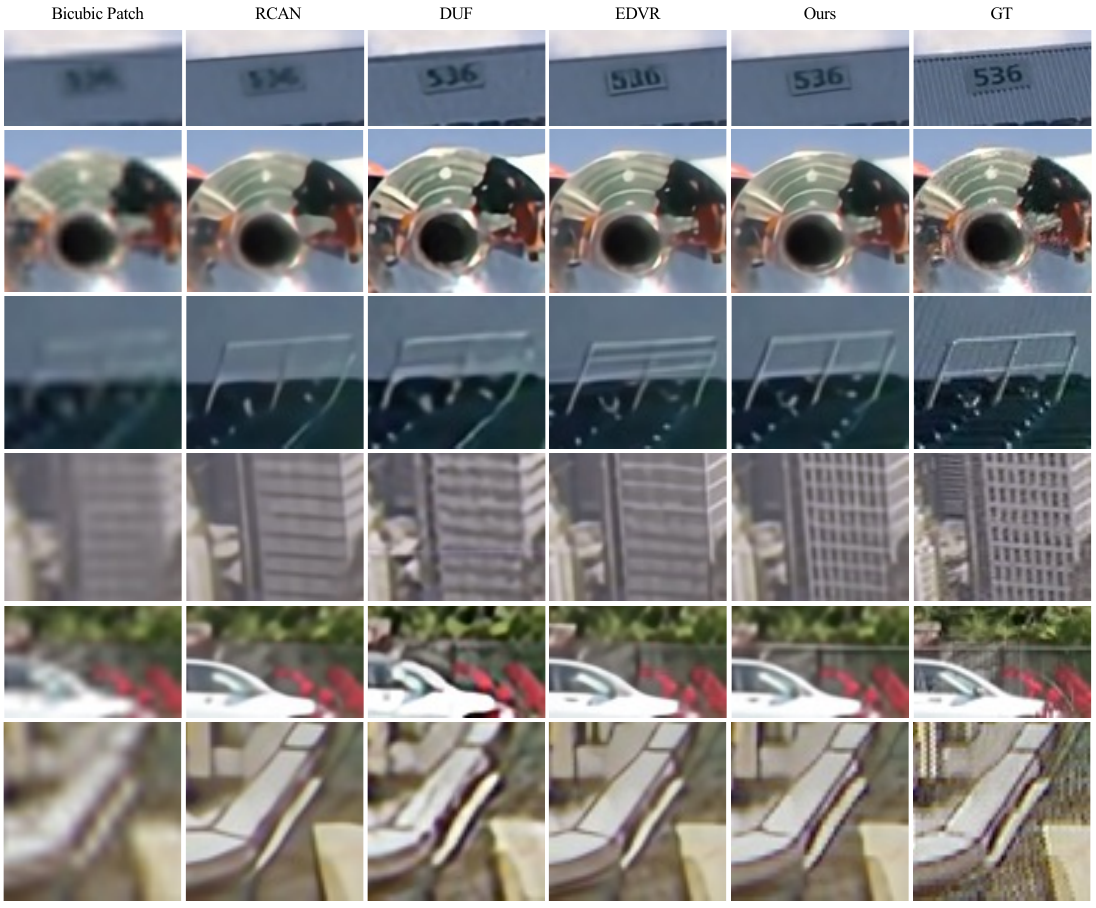}
	\caption{Qualitative comparison on UDM10~\cite{yi2019progressive} and Vid4~\cite{liu2013bayesian} for video SR.  } 
	\label{fig:vsr2}
\end{figure*} %

\begin{table*}[h]
	\begin{center}
		\footnotesize 
		\centering
		\begin{tabular}{l|c|c||c|c|c|c|c|c } 
			\hline
			Clip Name       & Bicubic                   &DUF~\cite{jo2018deep} &EDVR~\cite{wang2019edvr}  &MuCAN~\cite{li2020mucan}   &BasicVSR~\cite{chan2020basicvsr} &IconVSR~\cite{chan2020basicvsr}    &VSR-T~\cite{cao2021vsrt}    &Ours \\ 
			\hline
			Calendar (Y)    & 20.39/0.5720                                     &24.04/0.8110          &24.05/0.8147              &-                          & -     & -     &\textbf{{\color{blue}24.08}/{\color{blue}0.8125}}    &\textbf{{\color{red}24.65}/{\color{red}0.8270}}\\ 	
			City (Y)        & 25.16/0.6028                                  &\textbf{{\color{blue}28.27}/{\color{blue}0.8313}}          & 28.00/0.8122             &-                          & -      &- &27.94/0.8107                                &\textbf{{\color{red}29.92}/{\color{red}0.8428}}\\       
			Foliage (Y)     & 23.47/0.5666                                  &\textbf{{\color{red}26.41}/{\color{red}0.7709}}          &26.34/0.7635             &-                          & -         & -          &26.33/0.7635                      &\textbf{{\color{red}26.41}/{\color{blue}0.7652}}\\
			Walk (Y)        & 26.10/0.7974                                   &30.60/0.9141          &31.02/0.9152            &-                          & -         & -    &\textbf{{\color{blue}31.10}/{\color{blue}0.9163}}                            &\textbf{{\color{red}31.15}/{\color{red}0.9167}}\\  \hline 
			Average (Y)     & 23.78/0.6347                                     &27.33/\textbf{{\color{blue}0.8318}}          &27.35/0.8264                &27.26/0.8215      &27.24/0.8251                   & \textbf{{\color{blue}27.39}}/0.8279 &27.36/0.8258              &\textbf{{\color{red}27.90}/{\color{red}0.8380}} \\  \hline 
			Average (RGB)    & 22.37/0.6098                                      &25.79/\textbf{{\color{blue}0.8136}}          &\textbf{{\color{blue}25.83}}/0.8077              & -                         & -                               &-   &-       &\textbf{{\color{red}26.57}/{\color{red}0.8235}}      \\  \hline 
		\end{tabular}
	\end{center}
	\caption{Quantitative comparison on Vid4~\cite{liu2013bayesian} under x4 setting for video super-resolution. We report the PSNR~(dB)/SSIM results on both the RGB and the Y channel. Numbers in {\color {red} \bf red }and  {\color {blue} \bf blue} refer to the best and second-best results.}
	\label{table:vid4}
\end{table*}
\subsection{Comparison with State-of-the-art}
\begin{figure*}[h]
	\centering
	\includegraphics[width=1.0\linewidth]{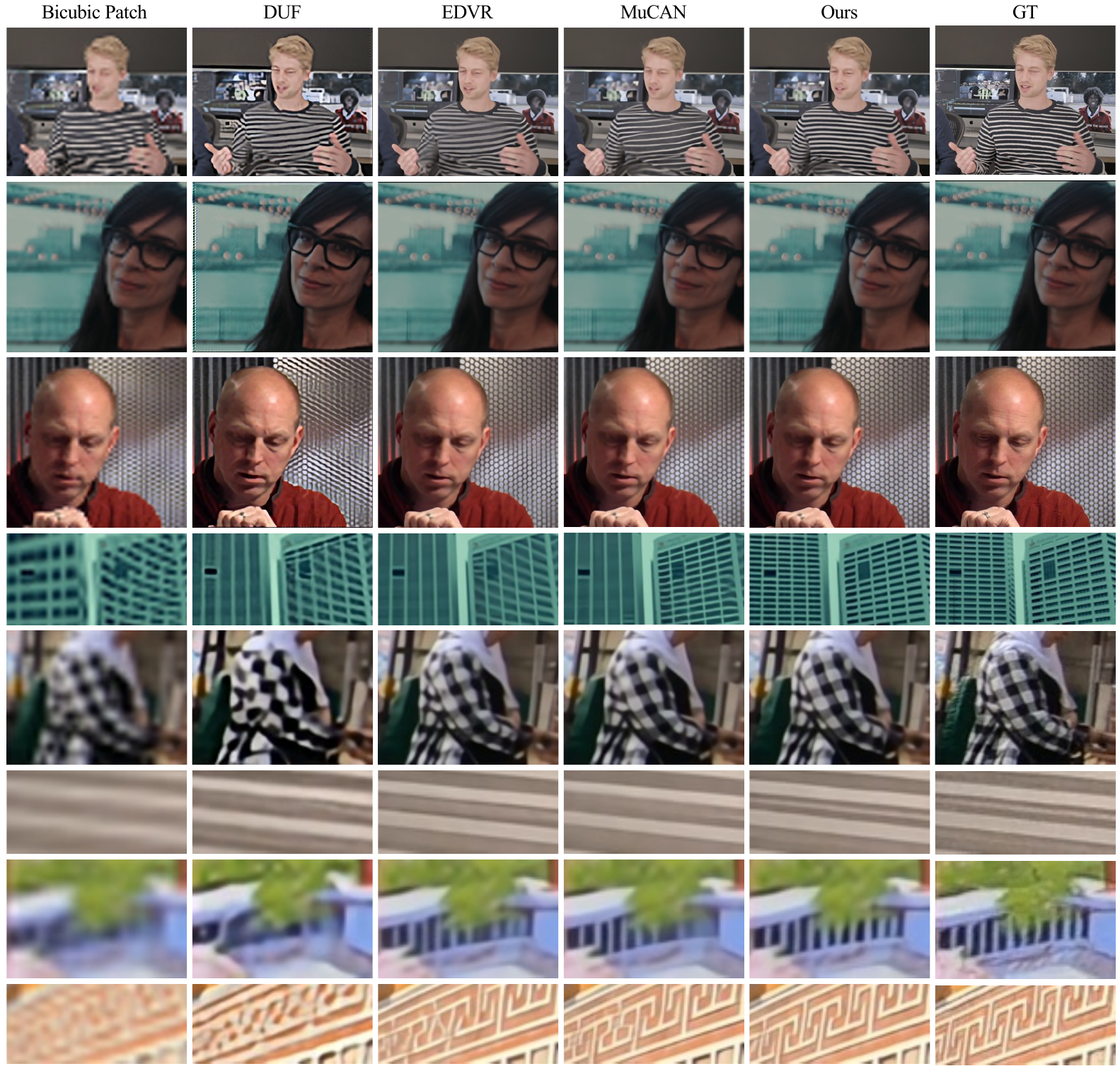}
	\caption{Qualitative comparison on Vimeo-90K-T~\cite{xue2019video} and REDS4~\cite{nah2019ntire} for video SR.  } 
	\label{fig:vsr1}
\end{figure*}

\begin{figure*}[t]
	\includegraphics[width=1.0\linewidth]{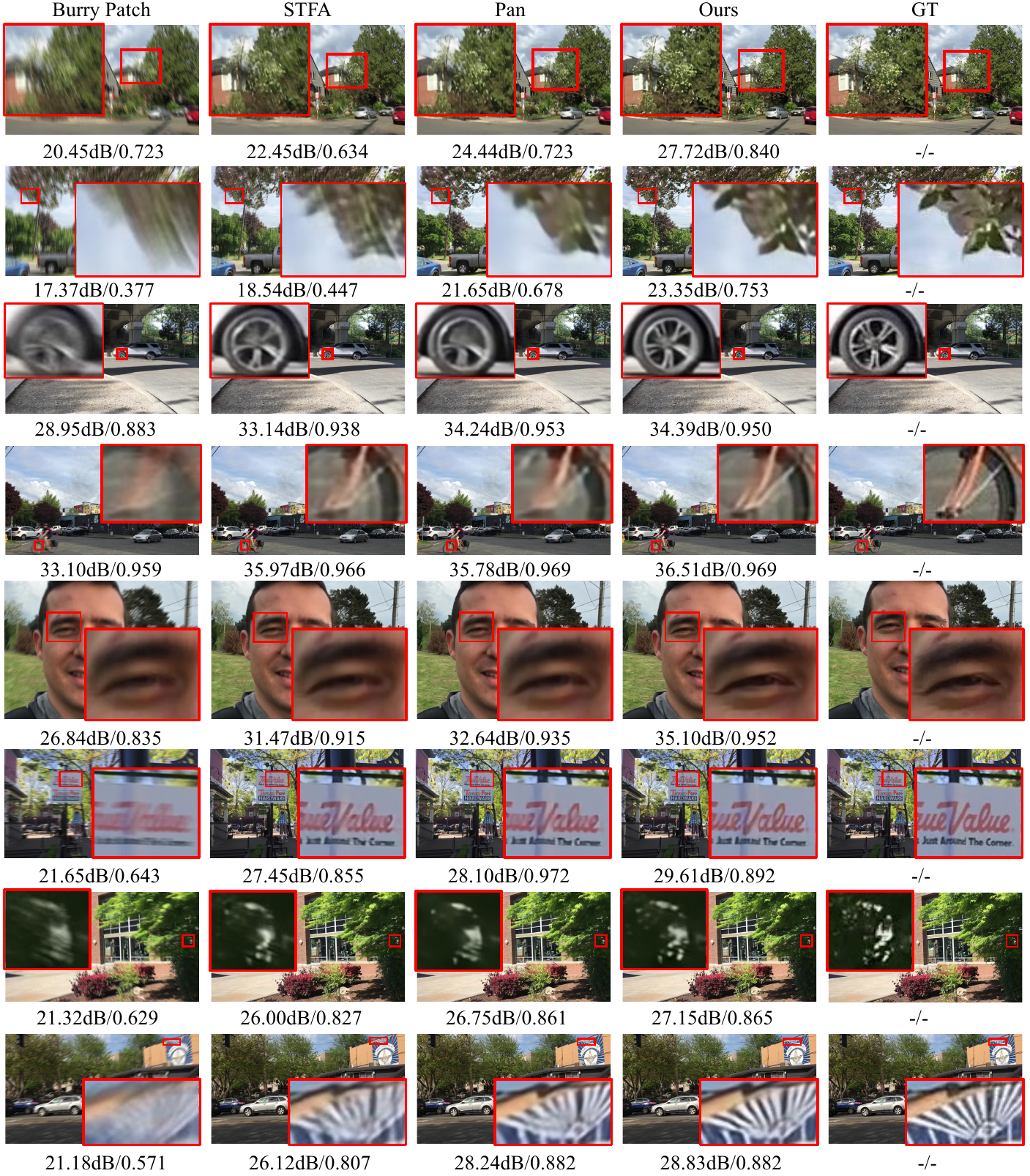}
	\caption{Qualitative comparison on VDB-T~\cite{su2017deep} for video deblurring.  } 
	\label{fig:deblur}
\end{figure*}

\begin{figure*}[t]
	\centering
	\includegraphics[width=1.0\linewidth]{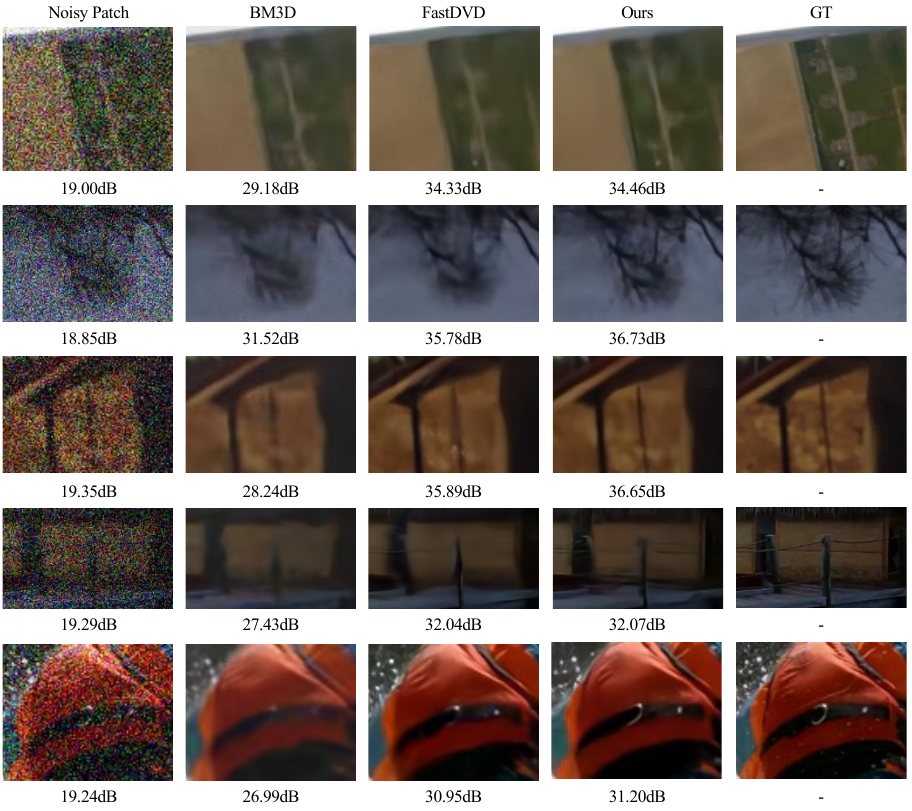}
	\caption{Qualitative comparison on Set8~\cite{tassano2020fastdvdnet}, DAIVS~\cite{khoreva2018video} for video denoising. The values beneath images represent the PSNR~(dB).  } 
	\label{fig:denoise}
\end{figure*}

In Table~\ref{table:VSR_SOTA_REDS} and~\ref{table:vid4}, we give the detialed comparison with several state-of-the-art video SR approaches~\cite{wang2019edvr,chan2020basicvsr,cao2021vsrt,cao2021vsrt} on REDS4~\cite{nah2019ntire} and Vid4~\cite{liu2013bayesian}. The PSNR and SSIM of each video sequence are reported. For most video clips of both two validation sets, our model \lly{consistently} achieves the best performance. Moreover, we provide \lly{extensive} qualitative comparison on UDM10~\cite{yi2019progressive}, Vid4~\cite{liu2013bayesian}, Vimeo-90K-T~\cite{xue2019video}, and REDS4~\cite{nah2019ntire} for video SR~(in Figure~\ref{fig:vsr2},~\ref{fig:vsr1}), VDB-T~\cite{su2017deep} for video deblurring~(in Figure~\ref{fig:deblur}) and Set8~\cite{tassano2020fastdvdnet}, DAIVS~\cite{khoreva2018video} for video denoising~(in Figure~\ref{fig:denoise}). All the qualitative results \lly{demonstrate} that our method \lly{has} the capacity to handle various challenging cases in \lly{these} three video restoration tasks.


\subsection{Video Results}
We also provide three videos for visual inspection.
{\bf ``city.mp4".   } This video illustrates the visual comparison between bicubic and our method on a Vid4 clip for video super-resolution. It can be observed that our method restores much clear image details~(e.g., the finer structure of buildings).\\
{\bf ``IMG0030.mp4".   } This video demonstrates the visual results of our method on a testing sequence of VDB-T~\cite{su2017deep} for the video deblurring task. The burry input and the generated frames are shown in it.\\
{\bf ``motorbike.mp4".   } \lly{This} video shows the restoration results on a sequence of Set8~\cite{tassano2020fastdvdnet} for video denoising.

	\end{appendices}
\end{document}